\documentclass[journal,comsoc]{IEEEtran}
\usepackage[T1]{fontenc}
\usepackage{cite}

\ifCLASSINFOpdf
   \usepackage[pdftex]{graphicx}
\else
   \usepackage[dvips]{graphicx}
\fi

\usepackage{bm}
\usepackage{amsmath}
\usepackage[cmintegrals]{newtxmath}

\usepackage[numbers]{natbib}
\usepackage{booktabs}
\usepackage{multirow}
\usepackage{multicol}
\usepackage{amsmath}

\usepackage{algorithm}
\usepackage{algorithmicx}
\usepackage{algpseudocode} 

\usepackage{mathtools}
\usepackage{hyperref}
\usepackage{enumitem}

\mathchardef\mhyphen="2D

\begin{document}
%
\title{A Layered Intuition–Method Model with Scope Extension for LLM Reasoning}

\author{Hong~Su
\IEEEcompsocitemizethanks{\IEEEcompsocthanksitem H. Su is with the School of Computer Science, Chengdu University of Information Technology, Chengdu, China.\\
 E-mail: suguest@126.com. \\
\protect\\
}
\thanks{}}

\markboth{Journal of \LaTeX\ Class Files,~Vol.~14, No.~8, August~2015}%
{Shell \MakeLowercase{\textit{et al.}}: Bare Demo of IEEEtran.cls for IEEE Communications Society Journals}
%

\maketitle

\begin{abstract}
Existing studies have introduced method-based reasoning and scope extension as approaches to enhance Large Language Model (LLM) performance beyond direct matrix mappings. Building on these foundations, this paper summarizes and integrates these ideas into a unified \textit{Intuition–Method Layered Model with Scope Extension}, designed to address indirected (unseen) issues more systematically. In this framework, intuition-based thinking provides rapid first-reaction answers, while method-based thinking decouples questions and solutions into transferable reasoning units. Scope extension is then applied to broaden applicability, including vertical (cause analysis), horizontal (parallel and generalized issues), and for the first time, \textit{temporal and spatial extensions}, which expand reasoning across time and contextual dimensions. 

These extensions are organized into systematic knowledge trees that interconnect into a knowledge network, thereby increasing adaptability. To quantitatively evaluate this process, we propose the \textit{entropy of method extension}, which measures the independence and diversity of extensions as an indicator of the system’s capacity to solve unseen questions. By logically connecting existing approaches with new extensions and introducing an entropy-based evaluation framework, this work advances toward a more robust and extensible reasoning paradigm for LLMs in real-world problem-solving.
\end{abstract}


\IEEEpeerreviewmaketitle

\section{Introduction}
Large Language Models (LLMs) have achieved remarkable performance across a wide range of natural language processing tasks \cite{chang2024survey}. Their success largely stems from the transformer architecture and massive pre-training, which enable strong mappings from input queries to output responses \cite{du2024stacking}. While this intuition-like mechanism allows LLMs to answer many direct questions, it remains inherently limited when confronted with indirected or unseen issues that lack explicit coverage in the pre-trained distribution. Such limitations restrict the ability of LLMs to adapt to complex real-world scenarios.

To address this gap, researchers have explored strategies such as \textit{method-based reasoning} \cite{su2025method}, which decouples questions and solutions into reusable methods \cite{zhang2025auragenome}, and \textit{scope extension} \cite{su2025improving}, which broadens the problem context to increase the chance of correct resolution. These approaches have demonstrated that reasoning can be extended beyond static matrix mappings \cite{zhao2025iam}. However, existing studies treat these ideas in isolation and do not provide a unified logical framework to connect them. Moreover, prior work has not systematically addressed how such extensions can be organized, nor has it provided quantitative measures of their effectiveness.

\begin{figure}[htp]
    \includegraphics[width=3.5in]{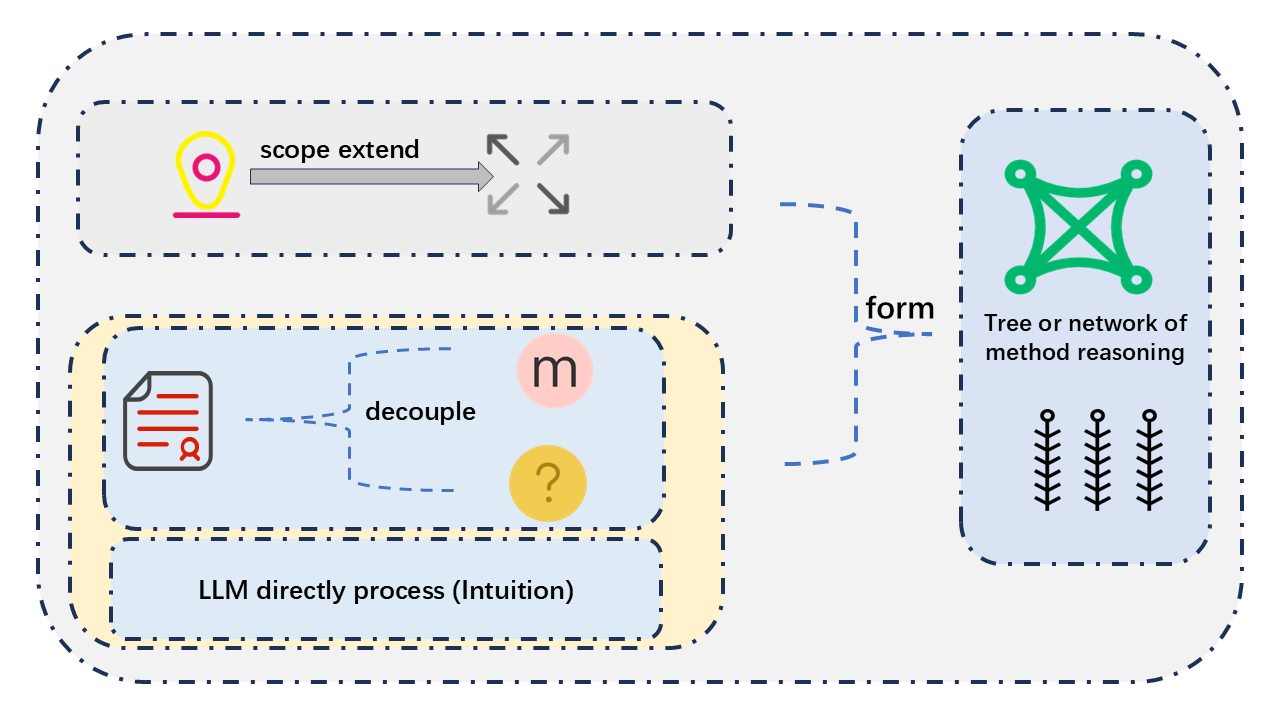}
    \caption{Intuition–Method Layered Model with Scope Extension for Forming Method (knowledge) Trees and Networks}
    \label{fig_intro}
\end{figure}

In this paper, we propose the \textit{Intuition–Method Layered Model with Scope Extension}, which integrates intuition-based reasoning, method-based reasoning, and multiple forms of scope extension into a unified framework. Intuition-based reasoning captures the direct, reflex-like outputs of LLMs, while method-based reasoning enables systematic reuse of question–solution pairs. Scope extension further improves adaptability by incorporating vertical (cause analysis) and horizontal (parallel and generalization) reasoning, as shown in Figure \ref{fig_intro}. Building on these foundations, this work makes the following key contributions:

\begin{itemize}
    \item We introduce the \textbf{method-layer architecture with scope extension}. While method-based reasoning and scope extension have been studied separately in prior work, we unify them into a single framework designed to address a broader range of indirected questions.  
    \item We propose \textbf{systematic knowledge trees}, which organize extensions into structured hierarchies and interconnect them into larger knowledge networks, enabling more systematic reasoning.  
    \item We define the \textbf{entropy of method extension}, a novel metric that quantifies the independence and diversity of extensions, providing a principled measure of an AI system’s capacity to solve unseen questions.  
\end{itemize}

By logically integrating existing techniques with these new contributions, our model extends the reasoning capacity of LLMs beyond pre-trained mappings. The framework enables more robust and systematic handling of indirected issues, moving toward an extensible reasoning paradigm that better supports real-world problem-solving.

The remainder of this paper is organized as follows. 
Section~\ref{sec_related_work} reviews related work on method-based reasoning, scope extension, difference-based reasoning, knowledge organization, and evaluation methods. 
Section~\ref{sec_think_model} presents the proposed Intuition–Method Layered Model with Scope Extension, which integrates intuition-based and method-based reasoning into a unified framework. 
Section~\ref{sec_scope_extension} further elaborates on scope extension strategies, including vertical, horizontal, temporal, and spatial dimensions, and their role in addressing unseen or indirected issues. 
Section~\ref{sec_Measurement} introduces the entropy of method extension as a quantitative framework for measuring reasoning diversity and adaptability. 
Finally, we conclude with a discussion of implications and future directions.

\section{Related Work} \label{sec_related_work}
The proposed model builds upon several strands of prior research that explore how Large Language Models (LLMs) can move beyond direct matrix-based mappings to handle more complex reasoning tasks. In particular, three areas are closely related to this study: (1) \textit{method-based reasoning}, which focuses on decoupling and reusing question–solution pairs; (2) \textit{scope extension approaches}, which expand the reasoning context to improve adaptability; and (3) \textit{difference-based reasoning}, which emphasizes recognizing temporal or spatial changes to guide decision-making.  

In addition, two complementary areas are also relevant: (4) \textit{knowledge organization in AI systems}, including knowledge graphs, memory architectures, and hierarchical reasoning structures; and (5) \textit{evaluation of reasoning diversity}, which highlights the limitations of existing benchmarks in measuring adaptability to unseen or indirected issues.  

This section reviews related works, identifies their limitations, and motivates the need for a unified framework that integrates method-based reasoning with scope extension. The novelty of this paper lies in unifying these approaches into a single framework capable of addressing a broader range of indirected questions, introducing systematic knowledge trees for dynamic organization, and proposing entropy-based evaluation to quantify reasoning diversity.

\subsection{Method-Based Reasoning in LLMs}
Traditional LLMs generate answers by relying on pre-trained transformer mappings between input and output tokens. While this mechanism works well for direct questions, it often fails to generalize when the question has not been explicitly seen during training. To address this limitation, researchers have proposed \textit{method-based reasoning}, which treats reasoning as the reuse of previously learned methods rather than as a one-time mapping.

A method is typically defined as a pair consisting of a question and its corresponding solution. By decoupling this pair, the method can be abstracted from its original context and reused for new but related questions. This approach enables LLMs to move beyond surface-level similarity and leverage transferable reasoning strategies \cite{su2025method}. Furthermore, recent work \cite{su2025cross} extends this idea by exploring how solutions can be applied to less similar questions within a method-based framework.

However, existing studies typically examine method reuse in isolation, without integrating it into a broader reasoning framework. Furthermore, they lack mechanisms for systematically extending methods to unseen issues, limiting their applicability to real-world problem-solving. In this paper, we build upon the idea of method-based reasoning and connect it with scope extension strategies, systematic knowledge structures, and entropy-based evaluation to form a more comprehensive model.

\subsection{Scope Extension Approaches}
Another line of research relevant to this work is \textit{scope extension}, which aims to broaden the reasoning context of LLMs when direct answers are insufficient. Instead of relying solely on the original input, scope extension incorporates additional information or perspectives to improve problem-solving ability.  

\subsubsection{Vertical Extension: Error Analysis and Causal Reasoning}
Vertical extension refers to identifying and incorporating causal factors related to a given problem. Prior studies have explored prompting LLMs to analyze their own errors or to incorporate explanatory reasoning \cite{su2025improving}. By including causal chains, models can correct earlier mistakes or provide more accurate answers. However, such approaches are often task-specific and lack a generalizable structure.

\subsubsection{Horizontal Extension: Parallel Reasoning and Generalization}
Horizontal extension focuses on exploring parallel or related issues. For example, when a direct solution fails, reasoning about related subproblems or generalizing from specific instances to broader categories can provide more robust answers \cite{su2025improving}. This technique improves adaptability but may also introduce noise, as not all parallel issues contribute useful context.

\subsubsection{Limitations of Existing Scope Extension Methods}
Although vertical and horizontal extensions have been proposed, prior work typically applies them in isolation. There is no unified framework that organizes different types of extensions or connects them systematically with method-based reasoning. Moreover, existing approaches rarely consider other dimensions of extension, such as temporal changes or spatial context, which are particularly important for handling dynamic real-world issues. These gaps motivate our proposal of temporal and spatial extensions, as well as the integration of all extensions into structured knowledge trees.

\subsection{Difference-Based Reasoning and Contextual Awareness}
In addition to method-based reasoning and scope extension, another important direction in LLM research is \textit{difference-based reasoning}, which emphasizes recognizing and responding to changes in context. Unlike static input–output mappings \cite{zhao2025iam}, this approach enables models to focus on what has changed across time or space, thereby improving their ability to handle evolving or indirected problems.  

\subsubsection{Temporal Difference Reasoning}
Several works have highlighted the importance of incorporating temporal differences into reasoning \cite{su2025difference}. For instance, monitoring how an environment or input sequence evolves over time allows models to identify critical changes that influence decision-making. Temporal reasoning has been applied in areas such as dialogue systems, forecasting, and process tracking, where historical context is essential.

\subsubsection{Spatial Reasoning and Attention Mechanisms}
Spatial difference reasoning focuses on identifying changes or salient features within visual or structured spatial data. Techniques such as attention mechanisms and region-based analysis \cite{su2025difference} have been developed to allow models to prioritize relevant parts of an input image or scene. These methods have improved performance in tasks like object detection and scene understanding by enabling localized reasoning.

\subsubsection{Remaining Challenges in Handling Indirected Questions}
While temporal and spatial difference reasoning has been partially explored, existing approaches are typically designed for specific applications (e.g., vision or sequence modeling) rather than as general reasoning strategies. Moreover, they are rarely integrated with method-based or scope-extended reasoning frameworks. As a result, prior methods lack the systematic capability to generalize across unseen or indirected questions. In this paper, we extend these ideas by formalizing temporal and spatial extensions within a unified layered framework, enabling broader adaptability in real-world scenarios.

\subsection{Knowledge Organization in AI Systems}
A related area of research focuses on how knowledge can be organized, represented, and retrieved in AI systems. Since LLMs are limited by the implicit structure of their pre-training data, explicit forms of knowledge organization have been proposed to improve reasoning, interpretability, and adaptability.  

\subsubsection{Knowledge Graphs and Structured Memory Approaches}
Knowledge graphs provide explicit representations of entities and their relationships, enabling structured reasoning across domains \cite{wang2017knowledge}. Similarly, structured memory architectures \cite{yogatama2018memory} allow models to store, update, and retrieve knowledge during interaction. These approaches improve factual consistency and enable systematic retrieval of related information. However, they are often static and require extensive manual or external construction.  

\subsubsection{Hierarchical Reasoning Networks}
Another stream of work investigates hierarchical reasoning structures, where knowledge is organized into layered or tree-like forms to support abstraction and decomposition \cite{zhang2025hierarchical}. These systems mimic aspects of human cognition by enabling reasoning at different levels of granularity. While effective in narrow domains, such frameworks are typically handcrafted and lack general adaptability to unseen problems.  

\subsubsection{Limitations Compared to Systematic Knowledge Trees}
Although knowledge graphs, memory modules, and hierarchical reasoning networks provide structured representations, they differ fundamentally from the \textit{systematic knowledge trees} proposed in this paper. Existing approaches generally represent static knowledge or domain-specific reasoning, while systematic knowledge trees are constructed dynamically through extensions (vertical, horizontal, temporal, and spatial). Moreover, knowledge trees in our framework interconnect to form a broader \textit{systematic knowledge network}, which increases reasoning diversity and adaptability. This distinction marks a key novelty of our work, as it introduces a dynamic, extension-driven organization of knowledge rather than relying on fixed structures.

\subsection{Evaluation of Reasoning Diversity}
Evaluating the reasoning ability of LLMs remains a significant challenge. Most existing benchmarks and metrics focus on task-specific accuracy or output quality, such as exact match, BLEU scores, or human evaluation \cite{celikyilmaz2020evaluation}. While these metrics are useful for measuring correctness in direct question answering, they do not capture the diversity or adaptability of reasoning strategies required for indirected or unseen problems.  

\subsubsection{Existing Performance Metrics}
Previous studies have introduced evaluation methods tailored to specific reasoning tasks, such as logical consistency checks \cite{clark2020transformers}, factual verification benchmarks \cite{guo2022survey}, or chain-of-thought quality assessments \cite{wei2022chain}. These approaches measure reasoning within narrowly defined settings, but they fail to account for how effectively a system can extend its reasoning to new contexts.  

\subsubsection{Lack of Metrics for Extension Independence and Diversity}
A critical limitation of current evaluation methods is the absence of a metric for measuring the \textit{independence} and \textit{diversity} of reasoning extensions. For example, two extensions that are tightly coupled add little new problem-solving capacity, while independent extensions significantly expand the system’s adaptability. Without a measure of this independence, it is difficult to quantify how well a model can generalize to indirected questions.  

\subsubsection{Motivation for Entropy of Method Extension}
To address this gap, we introduce the concept of \textit{entropy of method extension}. Unlike prior metrics, entropy provides a principled way to evaluate the diversity of reasoning strategies by quantifying the independence among extensions. Higher entropy corresponds to greater adaptability, reflecting the ability of the system to explore multiple perspectives and solve unseen problems. This contribution distinguishes our work from existing evaluation methods, as it shifts the focus from correctness alone to the extensibility and robustness of reasoning.

\section{Intuition Method Layered Model with Scope Extension to Solve Indirected Issues} \label{sec_think_model}

\subsection{Overview and Motivation}
Large Language Models (LLMs) primarily rely on the pre-trained transformer architecture to generate responses. While this matrix-based mapping is effective for many direct questions, it remains relatively fixed and limited when addressing unseen or indirected issues. To overcome these limitations, we propose the \textit{Intuition Method Layered Model with Scope Extension}. 

The central motivation of this model is to maximize the reuse of learned methods rather than relying solely on static parameter mappings. Instead of treating each question as an isolated case, questions are decomposed into two components: the \textit{question} and its corresponding \textit{solution} \cite{su2025method}. This decoupling enables methods to be systematically applied to a broader range of problems, even those not directly covered by the pre-trained model.

Our ultimate goal is to construct a structured system of reusable methods, analogous to how scientific disciplines (e.g., chemistry) organize knowledge into coherent frameworks. By forming such a systematic method network, an LLM can generalize beyond its training distribution, extend its reasoning scope, and solve a wider variety of real-world problems.

\subsection{Intuition and Method-Based Thinking Model}
Questions can be categorized into two types: \textit{direct} and \textit{indirected}. Direct questions are those for which an LLM can provide an immediate answer by relying on its pre-trained matrix. This process, which we call the \textit{intuition-based approach}, represents a first reaction to external input. Intuition arises from repeated training and functions as a natural reflex, but it is often limited in scope and rarely adaptable to new or complex situations. 

In contrast, \textit{indirected questions} lack a direct mapping in the pre-trained model. To address such cases, we employ a \textit{method-based approach}. A method is formally defined as a pair consisting of a \textit{question} and its corresponding \textit{solution}. By decoupling this pair, the method becomes a transferable unit of reasoning that can be applied to related questions beyond its original context \cite{su2025method}. 

This distinction leads to the formulation of the \textit{Intuition and Method-Based Thinking Model}. In this model, intuition serves as a fast but narrow mechanism for solving well-covered problems, while method-based reasoning provides a more systematic and extensible process. By combining the two, the model not only simulates human-like first reactions but also incorporates deliberate reasoning, thereby enhancing the ability to resolve indirected and unseen issues. 


\subsubsection{Timely Active Method-Based Thinking}
In addition to intuition and static method reuse, an effective AI system should engage in continuous and proactive reasoning, similar to human goal-directed thinking. We define this process as \textit{timely active method-based thinking}. 

In this approach, the system continuously monitors its current state and aligns its reasoning with a defined goal. The goal may be user-specified (e.g., through task input) or derived from the LLM’s output. For example, in autonomous driving, the overarching objective is safe driving, which remains constant while the environment dynamically changes.

To achieve this, the AI system actively collects contextual information and constructs prompts that integrate both the current state and the goal. The LLM is then queried for solutions that account for these factors. However, if the system directly queries without emphasizing recent or significant changes, the LLM may overlook critical details. To mitigate this, a \textit{difference-based prompting strategy} is proposed in\cite{su2025difference}: first prompting the LLM to identify key changes in temporal or spatial dimensions, and then requesting a solution that focuses on these changes. 

This mechanism simulates human-like active thinking, where attention is directed toward evolving aspects of the environment, thereby enabling more adaptive and timely decision-making.

\subsubsection{Method Improvement}
Beyond applying existing methods, an AI system should be capable of evaluating and refining them. This process, termed \textit{method improvement}, ensures that reasoning strategies remain effective and adaptable across diverse scenarios. 

After a method has been applied, the system can assess whether it achieved the intended outcome and whether alternative strategies might perform better. One direct approach is to prompt the LLM to critique the method as a whole, asking whether it can be improved. 

Another approach is to decompose the method into distinct stages and evaluate each step individually. By examining stages separately, the system can identify weaknesses, substitute improved steps, or revert unsuitable ones. This process is referred to as a \textit{step-change strategy}. 

Step-change strategies can be classified according to the degree of modification:
\begin{enumerate}[label=(\arabic*)]
    \item \textbf{Minimal change:} Only one step of the method is altered.  
    \item \textbf{Partial change:} A subset of steps is randomly or selectively modified.  
    \item \textbf{Complete change:} All steps are replaced, effectively generating a new method.  
\end{enumerate}

To measure the impact of these changes, two approaches are possible:  
(1) empirical validation in real-world or simulated environments, and  
(2) predictive evaluation by the LLM itself, particularly in high-risk contexts where direct testing is unsafe (e.g., highway driving scenarios).  

Through iterative refinement, method improvement enables continuous evolution of reasoning strategies, ensuring both robustness and adaptability in dynamic real-world environments.

\section{Scope Extension: Addressing Unseen or Indirected Issues} \label{sec_scope_extension}
While the method-based approach enables reasoning beyond direct mappings, certain questions may still remain unsolved or may yield incorrect answers. To broaden coverage, we introduce \textit{scope extension}, which expands the contextual boundaries of reasoning. Scope extension incorporates additional information, perspectives, and related issues, thereby enhancing the likelihood of producing correct or more complete answers \cite{su2025improving}.  

The major categories of scope extension include vertical, horizontal, temporal, and spatial dimensions. Together, these strategies enable systematic augmentation of the reasoning process.

\subsection{Vertical Extension (Cause Analysis) and Horizontal Extension (Parallel Issues and Generalization)}
Vertical extension \cite{su2025improving} focuses on identifying underlying reasons or causal factors. Errors in output can often be corrected by prompting the LLM to analyze the reasons behind its mistakes or by incorporating user-provided explanations. Formally, let $q$ denote the question and $y$ the predicted answer. A vertical extension introduces a cause variable $c$ such that:
\[
p(y \mid q) \;\;\longrightarrow\;\; p(y \mid q, c),
\]
where $c$ represents causal or explanatory information. By conditioning on $c$, the system reduces uncertainty in $y$ and improves problem resolution.  

Horizontal extension \cite{su2025improving} explores parallel or related issues that share contextual similarities. Let $\mathcal{N}(q)$ denote the set of neighboring or parallel questions related to $q$. Horizontal extension augments the reasoning context by:
\[
p(y \mid q) \;\;\longrightarrow\;\; p(y \mid q, \mathcal{N}(q)).
\]
This expansion increases the chance of retrieving relevant knowledge while mitigating misleading factors.  

A special case of horizontal extension is \textit{generalization}, where a mapping $g(q)$ transforms a specific question $q$ into a more general representation $q_g = g(q)$. If $M(q_g)$ denotes the set of methods applicable to $q_g$, then $M(q) \subseteq M(q_g)$. Thus, the generalized method offers broader transferability across multiple related problems.  

\subsection{Temporal and Spatial Extension}
In addition to reasoning extensions on the LLM side, the input data itself can also be extended. For instance, if an input image $X$ does not provide sufficient information, a larger region $X'$ can be included to support better judgment. This process is referred to as \textit{spatial extension}, defined as:
\[
X' = \mathcal{E}_{\mathrm{spatial}}(X), \quad X \subset X',
\]
where $\mathcal{E}_{\mathrm{spatial}}$ expands the spatial coverage of the input.  

Similarly, if knowledge of the history $H$ of an issue or predictions $F$ about its future evolution help to resolve it, we call this \textit{temporal extension}. In this case, the input sequence is extended from $X$ to:
\[
X' = X \cup H \cup F,
\]
where $H = \{x_{t-1}, x_{t-2}, \dots\}$ represents historical states and $F = \{x_{t+1}, x_{t+2}, \dots\}$ represents future predictions.  

\begin{figure}[htbp]
    \centering
    \includegraphics[width=3.5in]{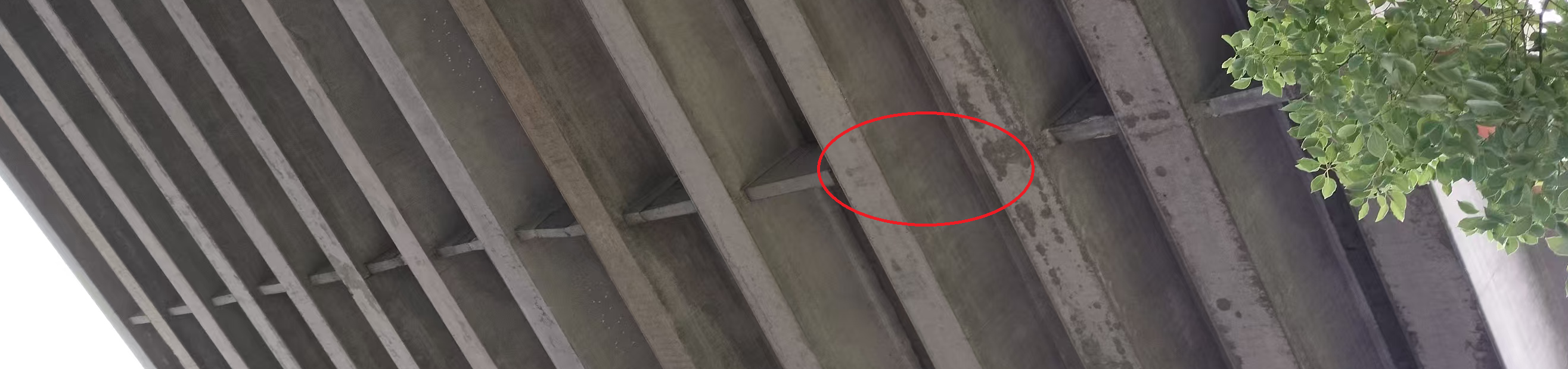}
    \caption{Initial reasoning on why the bridge lacks a connection, as indicated by the red circle}
    \label{fig_first}
\end{figure}

\begin{figure}[htbp]
    \centering
    \includegraphics[width=3.5 in]{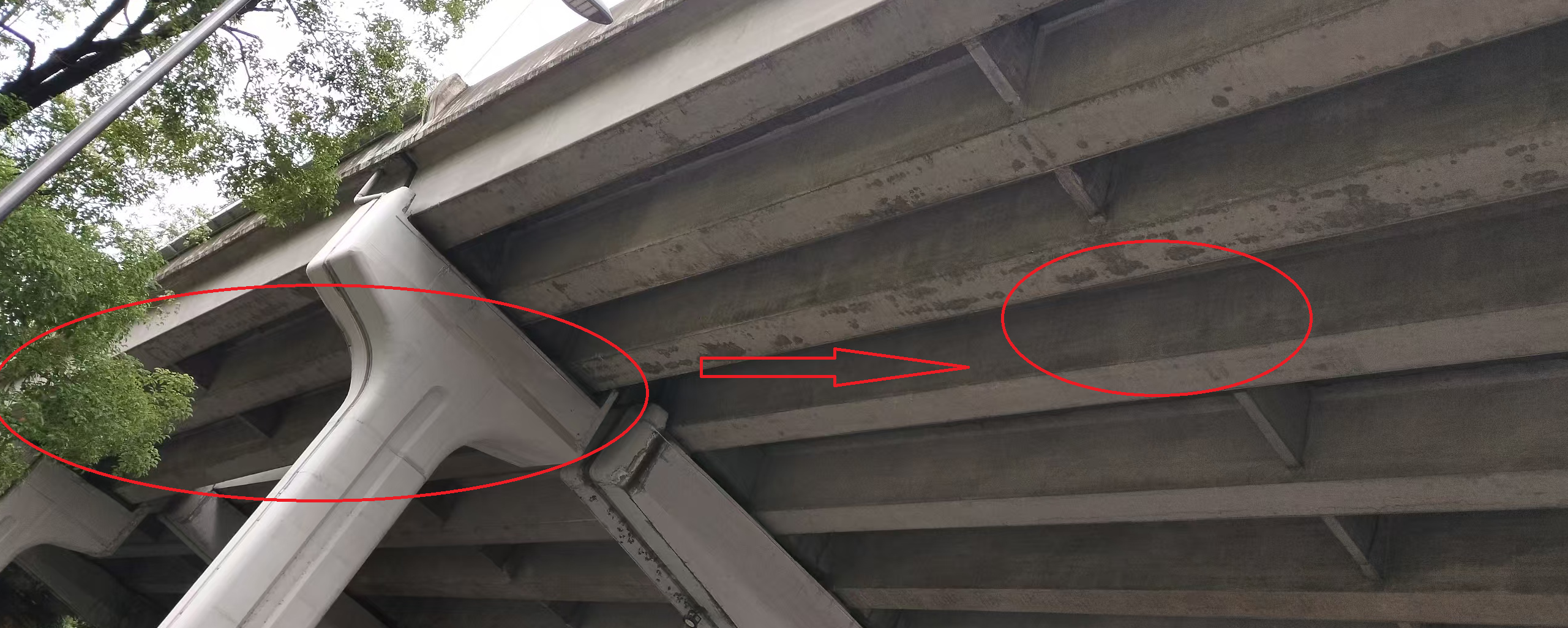}
    \caption{Extended reasoning with additional spatial context reveals the correct explanation: the road diverges at a fork to reduce one connection and minimize resonance.}
    \label{fig_extended}
\end{figure}

Consider the case of asking why a bridge lacks a connection. As illustrated in Fig.~\ref{fig_first}, an LLM may generate several possible reasons, many of which are incorrect. However, by extending the temporal or spatial scope—looking at a broader view of the bridge or considering its structural evolution—we may observe, as shown in Fig.~\ref{fig_extended}, that the bridge branches out but does not connect to the main road. This demonstrates how temporal and spatial extensions can reduce error by incorporating additional context.  

Another form of temporal and spatial extension is the \textit{scatter method} \cite{su2025scatter}. In this approach, an optimization identified at one stage or in one location is examined for applicability to other stages or locations. Formally, if an optimization $o$ is valid for stage $s_i$, the scatter method tests whether:
\[
o(s_i) \;\;\Rightarrow\;\; o(s_j), \quad \forall s_j \in \mathcal{S},
\]
where $\mathcal{S}$ is the set of alternative stages or locations. By transferring optimizations across time or space, the scatter method broadens the generalizability of solutions and enhances systematic reasoning.

\subsection{Completion of Scope Extension and Relation to LLM Output}
The extensions described above represent common and important strategies, but new forms of extension can be dynamically added. For instance, when interpreting a narrative, the emotional state of the author may serve as an additional dimension of scope. Thus, we maintain two lists: a \textit{common extension list} of standard strategies, and a \textit{dynamic extension list} for context-specific additions. Over time, frequently used dynamic extensions can be promoted into the common list. 

It is worth noting that LLM output itself can be seen as an implicit form of extension, since pre-trained models expand input text using vast background knowledge. Explicit scope extension, however, provides a rational and structured process for systematically broadening the reasoning context.

Formally, let $X$ denote the original input and $Y$ the output.  
\textbf{Implicit extension} occurs when the LLM internally augments $X$ with latent background knowledge $Z$ drawn from its pre-training corpus:
\[
p_\theta(Y \mid X) = \int p_\theta(Y \mid X, Z)\, q_\theta(Z \mid X)\, \mathrm{d}Z,
\]
where $q_\theta(Z \mid X)$ is the model’s implicit posterior over latent extensions.

\textbf{Explicit extension} instead applies a controlled operator $\mathcal{E}_{\mathrm{exp}}$ that enriches $X$ with structured external information $S = \{e_1, e_2, \dots, e_m\}$:
\[
X' = \mathcal{E}_{\mathrm{exp}}(X; S) = X \oplus \bigoplus_{e \in S} e,
\qquad
\hat{Y} = \arg\max_{y} p_\theta(y \mid X').
\]

The difference between implicit and explicit extension can be evaluated through measures such as KL-divergence:
\[
\mathrm{IG}(S \mid X) = \mathrm{KL}\!\left(p_\theta(\cdot \mid X') \,\big\|\, p_\theta(\cdot \mid X)\right),
\]
which quantifies the additional information contributed by explicit extensions beyond what is already encoded implicitly.

\subsection{Application of Methods to Dissimilar Questions}
Methods can naturally be reused for similar issues or through rational scope extension, where the relationship between questions is explicit. However, in many cases, a current question may lack direct similarity to any previously solved case. In such situations, it becomes necessary to \textit{borrow methods from dissimilar questions}.  

To enable this, we define a notion of \textit{distance} between questions. Let $q_i$ and $q_j$ denote two questions represented in an embedding space. Their similarity can be measured using cosine similarity:
\[
\text{sim}(q_i, q_j) = \frac{q_i \cdot q_j}{\|q_i\| \|q_j\|},
\]
and the corresponding distance is defined as:
\[
d(q_i, q_j) = 1 - \text{sim}(q_i, q_j).
\]
A smaller distance $d(q_i, q_j)$ indicates higher similarity and thus a greater likelihood that a borrowed method from $q_j$ will be effective for $q_i$. Conversely, larger distances may still provide alternative perspectives when no closer methods are available. In practice, multiple candidate methods may be borrowed simultaneously to increase coverage.  

This approach is referred to as \textit{distance-based method reuse}. Importantly, similarity need not be computed solely on the original question. By first applying scope extension to expand the problem context, similarity can be measured on the extended representation $q'_i$ instead of $q_i$. This often reveals meaningful connections between otherwise unrelated problems.  

Thus, when no direct method exists for a given question, the system proceeds in two steps:  
(1) apply scope extension to broaden the problem context, and  
(2) identify candidate methods to borrow based on distance to the extended question.  
In this way, distance-based reuse ensures that even dissimilar questions can be addressed by leveraging transferable reasoning strategies from other domains.

\subsection{Systematic Knowledge Formation}
The different forms of scope extension described above can be organized into structured representations of knowledge. Rather than treating each extension in isolation, we propose forming \textit{systematic knowledge trees}, which capture hierarchical and relational structures among questions, methods, and their extensions. When multiple trees are interconnected, they form a \textit{systematic knowledge network}.  

\subsubsection{Knowledge Trees from Extensions}
Each type of extension can be represented as a rooted tree $T = (V, E)$, where $V$ is the set of nodes (questions, methods, or contexts) and $E$ is the set of directed edges (extension relations). The structure of each tree depends on the type of extension:
\begin{itemize}
    \item \textbf{Horizontal extension:} siblings in $V$ represent parallel or related issues, while parent nodes correspond to generalized questions or methods. Formally, if $q$ is a specific question and $g(q)$ is its generalization, then $(g(q), q) \in E$.  
    \item \textbf{Vertical extension:} parent nodes capture causal factors, and child nodes represent their consequences. If $c$ is a cause and $e$ an effect, then $(c, e) \in E$.  
    \item \textbf{Temporal extension:} parent nodes represent historical states, and child nodes represent predicted or future states. For a sequence $\{x_{t-1}, x_t, x_{t+1}\}$, we have $(x_{t-1}, x_t), (x_t, x_{t+1}) \in E$.  
    \item \textbf{Spatial extension:} nodes correspond to contexts of different granularity. For example, a local region $r_{\text{local}}$ is connected to its global context $r_{\text{global}}$ by $(r_{\text{global}}, r_{\text{local}}) \in E$.  
\end{itemize}
Each tree is anchored to an original question or method, which serves as the root node $v_0 \in V$. These trees provide structured pathways for reasoning across multiple dimensions, systematically organizing strategies that would otherwise remain implicit.

\subsubsection{Knowledge Networks and Entropy of Extension}
When multiple knowledge trees share common nodes, they can be merged into a knowledge network $\mathcal{G} = (V, E)$, which is a directed acyclic graph (DAG). The union of trees increases coverage by connecting extensions across dimensions. For example, a parallel issue may also have a causal chain and a temporal trajectory, creating cross-tree links.  

Formally, if $T_1 = (V_1, E_1)$ and $T_2 = (V_2, E_2)$ are two trees with shared nodes $V_1 \cap V_2 \neq \emptyset$, the combined network is:
\[
\mathcal{G} = (V_1 \cup V_2, E_1 \cup E_2).
\]


This network-based organization enables the system to reach novel and previously unseen nodes, thereby increasing adaptability to indirected and real-world questions. By leveraging multiple independent extensions across interconnected trees, the model approaches the robustness and systematic reasoning typical of human knowledge.

\section{Measurement of Method Extension (Entropy Framework)} \label{sec_Measurement}
To evaluate the effectiveness of the proposed model, we introduce a quantitative framework based on the \textit{entropy of method extension}. The central idea is that an AI system’s reasoning capacity can be measured by the independence and diversity of its extensions. The more independent extensions the system can generate and apply, the greater its entropy and, consequently, its ability to handle unseen or indirected questions.

Formally, let $E = \{e_1, e_2, \dots, e_n\}$ represent the set of extensions applied to a question. Each extension $e_i$ contributes to the system’s reasoning space. The entropy of extension can then be defined as:
\[
H(E) = - \sum_{i=1}^{n} p(e_i) \log p(e_i),
\]
where $p(e_i)$ denotes the normalized contribution of extension $e_i$ to solving the question.  

Two key properties emerge:
\begin{itemize}
    \item \textbf{Coupled extensions:} If two extensions are strongly dependent, their joint contribution increases coverage only marginally, resulting in lower entropy gain.  
    \item \textbf{Independent extensions:} If two extensions address orthogonal aspects of a problem (e.g., temporal vs. causal reasoning), they contribute significant additional entropy.  
\end{itemize}

\paragraph*{Analysis of Method-Based Reasoning.}
Method-based reasoning also benefits from the entropy view. A single method $m$ reused across highly similar questions contributes relatively little entropy, as its scope of application is narrow. Formally, if $Q_m$ denotes the set of questions solvable by $m$, then
\[
H(Q_m) = - \sum_{q \in Q_m} p(q \mid m) \log p(q \mid m),
\]
remains low when $Q_m$ contains only closely related questions. However, when methods are decoupled from their original contexts and extended through independent transformations—such as generalization or distance-based reuse—the support of $Q_m$ expands, yielding higher entropy.  

The entropy gain of adding a new transformed method $m'$ can be expressed as:
\[
\Delta H = H(Q_m \cup Q_{m'}) - H(Q_m),
\]
where $Q_{m'}$ covers a distinct set of questions. If $Q_{m'}$ is largely independent of $Q_m$, then $\Delta H$ is significant, reflecting improved adaptability. Conversely, overlapping $Q_m$ and $Q_{m'}$ contribute little to overall entropy.  

In practice, the entropy measure thus distinguishes between systems that reuse a few tightly coupled methods (low entropy) and those that systematically apply diverse and independent methods (high entropy), with the latter reflecting stronger adaptability to unseen problems.

\paragraph*{Analysis of Scope Extension.}
Different forms of scope extension naturally map to this entropy framework. Vertical extensions often introduce causal variables $c$ that overlap with existing information, and thus add limited entropy when strongly correlated with the question $q$, i.e.,
\[
I(q;c) \approx H(c),
\]
indicating redundancy. In contrast, horizontal extensions expand the space of related or parallel issues $\mathcal{N}(q)$, contributing greater entropy when the distribution $p(\mathcal{N}(q))$ is diverse.  

Temporal and spatial extensions are typically orthogonal to vertical and horizontal ones. If two extensions $e_i$ and $e_j$ are independent, their joint contribution satisfies:
\[
H(e_i, e_j) = H(e_i) + H(e_j),
\]
thus yielding larger entropy gains by introducing new, non-overlapping dimensions of reasoning. This explains why combining multiple independent extensions leads to a more robust and adaptable problem-solving capacity.

\paragraph*{Knowledge Networks and Entropy.}
When extensions are organized into systematic knowledge trees and further interconnected into a knowledge network, the entropy of reasoning increases. Let $E_1, E_2, \dots, E_k$ represent the extension sets from $k$ different trees. The entropy of the combined network is
\[
H\!\left(\bigcup_{i=1}^k E_i\right) \;\;\geq\;\; \max_i H(E_i).
\]
If the extension sets are largely independent, then
\[
H\!\left(\bigcup_{i=1}^k E_i\right) \approx \sum_{i=1}^k H(E_i),
\]
indicating that the network contributes substantially more entropy than any single tree. By contrast, overlapping extensions provide limited additional entropy. Thus, the knowledge network systematically increases the system’s reasoning diversity, allowing it to handle a wider range of unseen problems.

Entropy thus provides a principled measure of how effectively an AI system broadens its reasoning scope. Higher entropy indicates that the system can solve a wider range of problems through diverse and independent extensions. This framework also provides a foundation for comparing different extension strategies, optimizing the balance between depth (improving methods) and breadth (extending scope).

\section{Conclusion}
This paper presented the \textit{Intuition–Method Layered Model with Scope Extension}, designed to enhance the reasoning capacity of Large Language Models (LLMs) when addressing indirected or unseen problems. We reviewed prior work on method-based reasoning and scope extension, and integrated them into a unified and logical framework. Building on these foundations, this work makes three main contributions. First, we propose the \textbf{method-layer architecture with scope extension}, which unifies method-based reasoning and scope extension into a single framework capable of addressing a broader range of indirected questions. Second, we introduce \textbf{systematic knowledge trees}, which organize extensions into hierarchical structures and interconnect them into larger knowledge networks. Third, we define the \textbf{entropy of method extension}, a novel metric that quantitatively measures the independence, diversity, and adaptability of reasoning strategies. Together, these contributions move LLMs beyond static matrix mappings toward a more systematic and extensible reasoning paradigm, thereby improving robustness in real-world problem solving.  

Future work will focus on further formalizing the mathematical foundations of the entropy framework and empirically validating its effectiveness across multiple benchmarks. In particular, we aim to quantify how different extensions (vertical, horizontal, temporal, and spatial) influence entropy and reasoning accuracy. Another promising direction is the automated construction of dynamic extension lists, enabling systems to evolve their repertoire of extensions over time. Finally, exploring efficient algorithms for building and traversing large-scale knowledge networks may facilitate practical deployment of the proposed framework in domains such as scientific discovery, autonomous systems, and complex decision support.

\ifCLASSOPTIONcaptionsoff
  \newpage
\fi

\bibliographystyle{IEEEtran}
\bibliography{ref}

%

\begin{IEEEbiography}{Hong Su}
  received the MS and PhD degrees, in 2006 and 2022, respectively, from Sichuan University, Chengdu, China. He is currently a researcher of Chengdu University of Information Technology Chengdu, China. His research interests include blockchain, cross-chain and smart contract.
\end{IEEEbiography}




\end{document}